\documentclass[a4paper]{svproc}
\usepackage{url}

\usepackage{hyperref}
\usepackage{subcaption}
\usepackage{tabularx}
\usepackage{wrapfig}
\newcolumntype{Y}{>{\centering\arraybackslash}X} 

\usepackage{comment}
\usepackage{siunitx}
\usepackage{relsize}
\usepackage{ifthen}
\usepackage[colorinlistoftodos]{todonotes}

\usepackage[vlined,ruled,linesnumbered]{algorithm2e}
\usepackage{graphics} 
\usepackage{rotating}
\usepackage{color}
\usepackage{enumerate}
\usepackage[T1]{fontenc}
\usepackage{psfrag}
\usepackage{epsfig} 
\usepackage{booktabs}
\usepackage{graphicx,url}
\usepackage{multirow}
\usepackage{array}
\usepackage{latexsym}
\usepackage{amsfonts}
\usepackage{amsmath}
\usepackage{amssymb}
\usepackage{xstring}
\usepackage[noend]{algorithmic}
\usepackage{multirow}
\usepackage{xcolor}
\usepackage{prettyref}
\usepackage{flexisym}
\usepackage{bigdelim}
\usepackage{breqn} 
\usepackage{listings}

\usepackage{enumitem}
\usepackage{xspace}
\usepackage{bm}
\graphicspath{{./figures/}}
\usepackage{tikz}
\usetikzlibrary{matrix,calc}

\usepackage{mdwlist}

\makecompactlist{itemize}{stditemize}

\newrefformat{prob}{Problem\,\ref{#1}}
\newrefformat{def}{Definition\,\ref{#1}}
\newrefformat{sec}{Section\,\ref{#1}}
\newrefformat{sub}{Section\,\ref{#1}}
\newrefformat{prop}{Proposition\,\ref{#1}}
\newrefformat{app}{Appendix\,\ref{#1}}
\newrefformat{alg}{Algorithm\,\ref{#1}}
\newrefformat{cor}{Corollary\,\ref{#1}}
\newrefformat{thm}{Theorem\,\ref{#1}}
\newrefformat{lem}{Lemma\,\ref{#1}}
\newrefformat{fig}{Fig.\,\ref{#1}}
\newrefformat{tab}{Table\,\ref{#1}}

\newcommand{\bdmath}{\begin{dmath}}
\newcommand{\edmath}{\end{dmath}}
\newcommand{\beq}{\begin{equation}}
\newcommand{\eeq}{\end{equation}}
\newcommand{\bdm}{\begin{displaymath}}
\newcommand{\edm}{\end{displaymath}}
\newcommand{\bea}{\begin{eqnarray}}
\newcommand{\eea}{\end{eqnarray}}
\newcommand{\beal}{\beq \begin{array}{ll}}
\newcommand{\eeal}{\end{array} \eeq}
\newcommand{\beas}{\begin{eqnarray*}}
\newcommand{\eeas}{\end{eqnarray*}}
\newcommand{\ba}{\begin{array}}
\newcommand{\ea}{\end{array}}
\newcommand{\bit}{\begin{itemize}}
\newcommand{\eit}{\end{itemize}}
\newcommand{\ben}{\begin{enumerate}}
\newcommand{\een}{\end{enumerate}}

\newcommand{\eg}{\emph{e.g.,}\xspace}
\newcommand{\ie}{\emph{i.e.,}\xspace}
\newcommand{\myParagraph}[1]{\textbf{#1.}\xspace}

\renewcommand{\boldsymbol}[1]{{\bm #1}}

\newcommand{\hiddenText}{{\color{gray} hidden text.}}
\newcommand{\hideWithText}[1]{\hiddenText}

\newcommand{\blue}[1]{{\color{blue}#1}}

\newcommand{\linkToPdf}[1]{\href{#1}{\blue{(pdf)}}}
\newcommand{\linkToPpt}[1]{\href{#1}{\blue{(ppt)}}}
\newcommand{\linkToCode}[1]{\href{#1}{\blue{(code)}}}
\newcommand{\linkToWeb}[1]{\href{#1}{\blue{(web)}}}
\newcommand{\linkToVideo}[1]{\href{#1}{\blue{(video)}}}
\newcommand{\linkToMedia}[1]{\href{#1}{\blue{(media)}}}
\newcommand{\award}[1]{\xspace} 

\newcommand{\bmat}{\left[ \begin{array}}
\newcommand{\emat}{\end{array} \right]}

\newcommand{\bal}{\begin{align}}
\newcommand{\eal}{\end{align}}

\newcommand{\specialcell}[2][c]{%
  \begin{tabular}[#1]{@{}c@{}}#2\end{tabular}}

\usepackage{xspace}

\begin{document}
\mainmatter              
%
%
\title{Box Pose and Shape Estimation and Domain Adaptation for Large-Scale Warehouse Automation}
\titlerunning{BOSS}  
%
\author{
  Xihang Yu\inst{1} \and 
  Rajat Talak\inst{1} \and 
  Jingnan Shi\inst{1} \and 
  Ulrich Viereck\inst{2} \and \\
  Igor Gilitschenski\inst{2,3} \and
  Luca Carlone\inst{1}
}
\authorrunning{Xihang Yu et al.} 
%
\tocauthor{
  Xihang Yu, 
  Rajat Talak, 
  Jingnan Shi, 
  Ulrich Viereck, 
  Igor Gilitschenski, 
  Luca Carlone
}
\institute{
  Laboratory for Information \& Decision Systems (LIDS) \\
  Massachusetts Institute of Technology, Cambridge, USA,\\
    \email{\{jimmyyu,talak,jnshi,lcarlone\}@mit.edu}
  \and
  Symbotic, Wilmington, Massachusetts,\\
    \email uviereck@symbotic.com
  \and
 Vector Institute \\
 University of Toronto, Ontario, Canada,\\
    \email gilitschenski@cs.toronto.edu
}
\maketitle              



\begin{abstract} 
Modern warehouse automation systems rely on fleets of intelligent robots that generate vast amounts of data --- most of which remains unannotated. This paper develops a self-supervised domain adaptation pipeline that leverages real-world, unlabeled data to improve perception models without requiring manual annotations. Our work focuses specifically on estimating the pose and shape of boxes and presents a correct-and-certify pipeline for self-supervised box pose and shape estimation. We extensively evaluate our approach across a range of simulated and real industrial settings, including adaptation to a large-scale real-world dataset of 50,000 images. The self-supervised model significantly outperforms models trained solely in simulation and shows substantial improvements over a zero-shot 3D bounding box estimation baseline.
\end{abstract}


%

\begin{figure*}[t]
    \centering
    \includegraphics[width= 1.0\columnwidth]{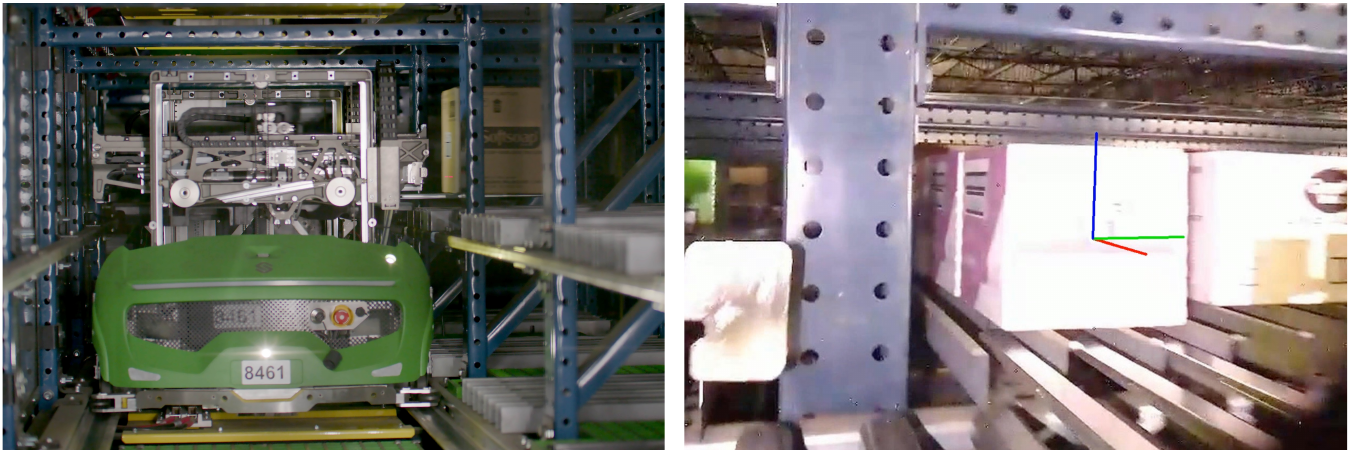}
    \caption{(left) Robot in a Symbotic warehouse picking up a box from a shelf. (right) A real-world pick-up task using a model trained entirely in simulation and adapted with our self-supervised pipeline on unlabeled data.}
    \label{fig:motivation}
\end{figure*}

\section{Introduction and Problem Statement}
\label{sec:introduction}

Warehouse automation has the potential to increase operational efficiency and accuracy while reducing labor costs and human errors. 
A key task in this process involves robots picking, transporting, and placing boxes between buffer and storage shelves (see~\autoref{fig:motivation}). 
Executing such tasks reliably over long durations without failure requires accurate perception in the operating domain.

In this work, we consider the problem of estimating the pose and shape of boxes encountered in warehouse automation applications. 
We parameterize the box as a cuboid and aim to simultaneously estimate its pose $\boldsymbol{T}\in\mathrm{SE}(3)$ and shape $\boldsymbol{S}$ (\ie width, height, and depth).
Automated warehouses are a source of large amounts of unannotated data, collected by the robots during operation. 
\textit{Our aim is to use the large-scale unannotated data collected by robots and enable self-supervised domain adaptation to improve the perception results.}

\myParagraph{Contributions}
Our contributions are threefold: (1) We propose a pipeline that can accurately estimate the pose and shape of a box from stereo images. (2) We implement a self-training pipeline, leveraging the \textbf{correct-and-certify} approach from ~\cite{yang2021self,shi2023correct,talak2023certifiable,shi2024crisp,jawaid2024test}. The approach utilizes corrected and certified estimates to self-train the model and avoids the need for data annotation. (3) We report an industry-scale demonstration of accurate box pose and shape estimation in the desired operating domain. This is made possible by self-training on a dataset of 50,000 images collected from \href{https://www.symbotic.com/}{Symbotic} warehouses.


\section{Related Work}
\label{sec:relatedWork}

\subsection{Category-Level Object Pose and Shape Estimation}

Object pose and shape estimation involves recovering the 3D pose and shape of an object. Existing methods can be classified based on whether they assume access to known instance-level shape priors. Approaches that rely on known shape priors typically use predefined CAD models of each object instance \cite{talak2023certifiable,shi2023correct}. In contrast, category-level methods aim to generalize across unseen instances within the same object category, without requiring instance-specific CAD models. These approaches often learn to model shape deformations or normalized coordinate representations to capture intra-class variation \cite{pavlakos20176,wang2019normalized,tian2020shape,fu2022category,shi2024crisp}.

In this work, we focus on estimating the pose and shape of box-like objects without relying on instance-level shape priors. Several prior methods have addressed this problem from different perspectives. For example, \cite{kai2025pallet} proposes Front Face Shot (FFS), a method that estimates box pose from front-view RGB images. While FFS generalizes well to unseen pallet appearances, it depends heavily on accurate front-face visibility and bounding box detection, which limits its robustness in the presence of occlusion. Another approach, Cube R-CNN \cite{brazil2023omni3d}, is a zero-shot RGB-only method trained on the large-scale Omni3D benchmark for general 3D bounding box prediction. However, in our experiments, it suffers from substantial performance degradation due to domain shift, making it less effective for our target setting.

\subsection{Test-Time Adaptation}

Test-time adaptation has been explored through various strategies. Related works \cite{sun2019unsupervised,sun2020test} leverage auxiliary tasks, \eg image rotation prediction, to guide feature learning during test time.~\cite{hansen2020self} generalizes this idea to reinforcement learning, where action-observation pairs naturally serve as feedback signals. Another line of work focuses on domain-level consistency across a mini-batch of test inputs by minimizing softmax-entropy loss at test time~\cite{wang2020tent,goyal2022test}. To handle the more challenging scenario of having only a single test sample,~\cite{zhang2022memo} uses data augmentation to synthesize a mini-batch. Temporal consistency has also been leveraged as a source of self-supervision~\cite{zurbrugg2022embodied,merrill2022symmetry}. These methods maintain a coherent 3D scene over time and render it into 2D views to provide consistent supervisory signals for 2D vision tasks.

Another stream of methods follows a correct-and-certify paradigm~\cite{yang2021self,shi2023correct,talak2023certifiable,shi2024crisp,jawaid2024test}, where model outputs are first corrected, and only those that pass certain certification criteria are used as pseudo-labels for self-training. These methods often rely on auxiliary sensor inputs such as CAD models~\cite{jawaid2024test}, depth maps~\cite{shi2024crisp}, or segmentation masks~\cite{shi2023correct}. In contrast, our approach does not assume additional sensor modalities. Instead, it relies on the SAM2 model~\cite{ravi2024sam2}, making the framework simple and easily adaptable to a variety of warehouse automation tasks. 


\section{Technical Approach}
\label{sec:problemStatement}

\begin{figure*}[t]
    \centering
    \includegraphics[width= 1.0\columnwidth]{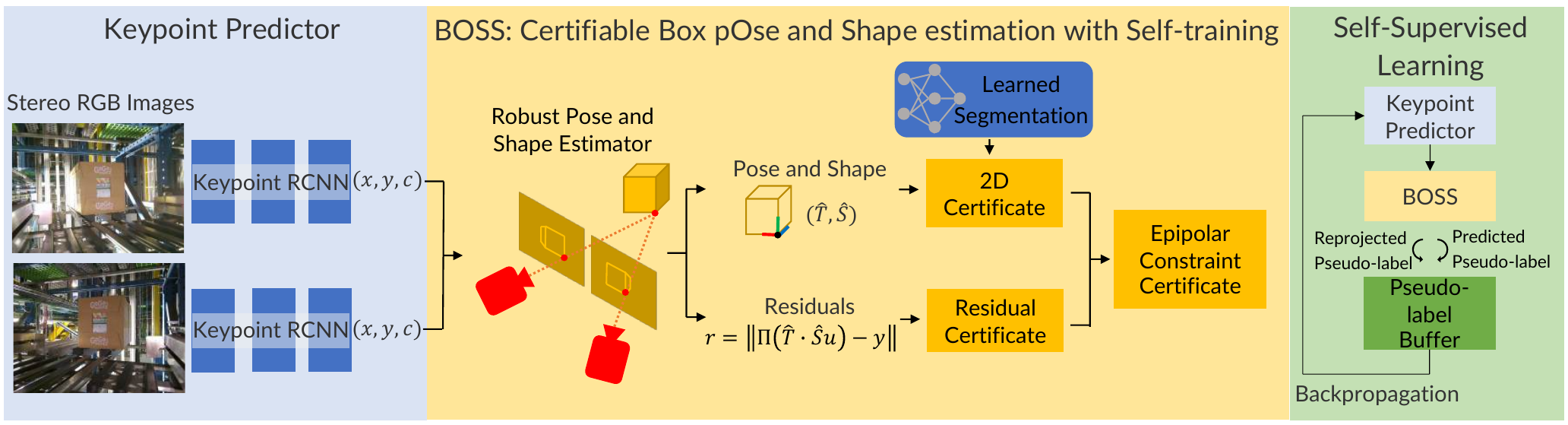}
    \caption{Illustration of the proposed pipeline. We take stereo images as input and use two keypoint prediction networks --- one for each view --- to predict the box corners. Only high-confidence confidence keypoints are used for pose and shape estimation. The box pose and shape estimation problem is formulated as a two-view Perspective-n-Point optimization. Then, pose and shape estimates that pass certain checks are used to generate pseudo-labels for self-supervised learning. In particular, a predicted keypoint is considered a valid pseudo-label for self-supervised learning if it passes a number of image-level and keypoint-level checks (certificates). To ensure robustness against outliers, we apply Geman-McClure \cite{barron2019general} as a robust loss in the pose and shape estimator.}
    \label{pipeline}
\end{figure*}

We consider a robot operating in a warehouse environment, equipped with two calibrated RGB stereo-cameras. These cameras capture RGB images of 3D scenes that contain an object of interest. We assume the object to be a storage box parametrized by its width, height, and depth $(a,b,c)$. Let \( \boldsymbol{S} = \text{diag}(a,b,c) \) represent the anisotropic scaling factors to a unit cube (\ie a cuboid with all edges of length 1). The goal is to compute the pose and shape of the box.~\autoref{pipeline} shows our pipeline. 
It consists of a pre-trained stereo keypoint detection model trained on the small labeled dataset or in simulation, 
an estimator to compute the pose $\boldsymbol{T}$ and the shape $\boldsymbol{S}$ of the box in the 3D scene, and a self-training procedure for the keypoint detection model to improve pose estimation on unlabelled data. We call the resulting approach \texttt{BOSS} (Box pOse and Shape estimation with Self-training).

\paragraph{Keypoint Predictor.} We use Keypoint-RCNN as our keypoint predictor network~\cite{he2017mask}, with one network for each view. The network outputs the corners of the boxes as keypoints with confidence scores. 
We only keep keypoints with a confidence score greater than a specified threshold $\epsilon_{conf}$. 
\paragraph{Stereo Box Pose and Shape Estimator.}
We estimate the box pose and shape through a two-view Perspective-n-Point (PnP) optimization problem.
The objective is to estimate the object pose and scaling factors that align the reprojected keypoints with their predicted keypoints in the two camera views. The optimization problem is formulated as:
\beq
\min_{\boldsymbol{T}, \boldsymbol{S}} \left\{ \sum_{i=1}^{N^l} \left\| \boldsymbol{\delta^l_i} \right\|_2^2 
+ \sum_{i=1}^{N^r} \left\| \boldsymbol{\delta^r_i} \right\|_2^2 \right\},
\label{eq:optimization}
\eeq
where $
\boldsymbol{\delta^l_i} = \Pi^l(\boldsymbol{T} \cdot \boldsymbol{S} \boldsymbol{u}_i) - \tilde{\boldsymbol{y}}^l_i$ and $\boldsymbol{\delta^r_i} = \Pi^r \big({\boldsymbol{T}}^r_l \cdot \boldsymbol{T} \cdot \boldsymbol{S} \boldsymbol{u}_i\big) - \tilde{\boldsymbol{y}}^r_i$ are the distances between the $i^{th}$ reprojected and predicted keypoints. \( N^l \) and \( N^r \) are the numbers of keypoints in the left and right images, respectively, \(  \boldsymbol{u}_i \) are the 3D keypoints on the unit cube centered at the origin of the object frame, \( \tilde{\boldsymbol{y}}^l_i \) and \( \tilde{\boldsymbol{y}}^r_i \) are the observed 2D keypoint positions in the left and right images, \( {\boldsymbol{T}}^r_l\) is the known transformation from the left to the right camera frame, \( \Pi^l \) and \( \Pi^r \) represent the projection models of the left and right cameras, respectively. In words, Equation~\eqref{eq:optimization} minimizes the mismatch between the projections of the estimated box corners (parametrized by the pose $\boldsymbol{T}$ and shape $\boldsymbol{S}$) and the keypoint measurements. The optimization is solved via gradient descent in PyTorch \cite{paszke2019pytorch}. We relax the rotation matrix constraint and, in each iteration, project the optimized rotation back onto \( \text{SO}(3) \) using SVD.

\paragraph{Self-Training.} 
To self-train, we use certificates to select pseudo-labels. 
We use a 2D certificate, a residual certificate, and an epipolar constraint certificate. We admit keypoints as pseudo-labels only if they pass all three certificates.

Let $\hat{\boldsymbol{T}}$ and $\hat{\boldsymbol{S}}$ be the estimated pose and shape from Equation~\eqref{eq:optimization} respectively. Our 2D certificate is based on intersection over union (IoU), given by
\beq
\mathcal{OC}_{2D}(\hat{\boldsymbol{S}}, \hat{\boldsymbol{T}}) = 
\mathbb{I}\left\{ 
\frac{\text{ar}(\boldsymbol{M} \cap \hat{\boldsymbol{M}})}{\text{ar}(\boldsymbol{M} \cup \hat{\boldsymbol{M}})} > 1 - \epsilon_{2D}
\right\},
\label{eq:2d}
\eeq
where $\text{ar}(\boldsymbol{M})$ denotes the pixel area of all pixels $(i, j)$ in the mask $\boldsymbol{M}$ with $\boldsymbol{M}(i, j) = 1$, and $\epsilon_{2D}$ is a given threshold. IoU is computed using a reprojected 3D model $\hat{\boldsymbol{M}}$ and ground truth (GT) or detected segmentation mask $\boldsymbol{M}$.

Our residual certificate filters keypoints based on residuals in Equation~\eqref{eq:optimization}. Let $\boldsymbol{\delta}$ represent the residuals in the pose and shape estimator, indexed as $\boldsymbol{\delta}^l$ and $\boldsymbol{\delta}^r$ for the left and right views, respectively. The residual-based certificate is defined as
\beq
\mathcal{OC}_{res}(\tilde{\boldsymbol{y}}, \bar{\boldsymbol{T}}) = 
\mathbb{I}\left\{ 
\left\|\boldsymbol{\delta} \right\|_2 < \epsilon_{res}
\right\}.
\label{eq:res}
\eeq
where $\left\| \cdot \right\|_2$ is $l^2$-norm and $\epsilon_{res}$ is a tunable threshold. If $\mathcal{OC}_{res}(\tilde{\boldsymbol{y}}, \bar{\boldsymbol{T}}) = 1$, we use $\tilde{\boldsymbol{y}}$ as a pseudo-label. Otherwise, if $\mathcal{OC}_{res}(\tilde{\boldsymbol{y}}, \bar{\boldsymbol{T}}) = 0$, we use the reprojected keypoint $\Pi(\bar{\boldsymbol{T}} \cdot \hat{\boldsymbol{S}} \boldsymbol{u}_i)$ as a pseudo-label, where $\bar{\boldsymbol{T}} = \hat{\boldsymbol{T}}$ for the left view and $\bar{\boldsymbol{T}} = {\boldsymbol{T}}^r_l \cdot \hat{\boldsymbol{T}}$ for the right view.

We use an epipolar constraint certificate as a final check. Given the known intrinsics and extrinsics of both cameras, we rectify the keypoints so that epipolar lines align with the x-axis, ensuring corresponding points share the same y-coordinates. We then verify the consistency of these y-coordinates in the rectified frames. Denote the rectified keypoints $\tilde{\boldsymbol{y}}_l$ and $\tilde{\boldsymbol{y}}_r$ be $\tilde{\boldsymbol{y}}'_l$ and $\tilde{\boldsymbol{y}}'_r$, respectively. Then epipolar constraint certificate is defined as:
\beq
\mathcal{OC}_{epi}(\tilde{\boldsymbol{y}}_l, \tilde{\boldsymbol{y}}_r) = 
\mathbb{I}\left\{ 
(\tilde{\boldsymbol{y}}'_l - \tilde{\boldsymbol{y}}'_r)[2] < \epsilon_{epi}
\right\}.
\label{eq:epi}
\eeq
where $(\cdot)[2]$ denotes the y-coordinate and $\epsilon_{epi}$ is a given threshold. \\

\section{Certificate Validation}
\label{sec:certificate_validation}

In this section, we empirically validate the three certificates. Over an annotated dataset we show that the certificate scores correlate highly with the ground-truth metrics such as the keypoint root mean square error (RMSE). We compare the three certificate values: (i) IoU  (see Equation~\eqref{eq:2d}), (ii) $\left\|\boldsymbol{\delta} \right\|_2$ (see Equation~\eqref{eq:res}), and (iii) $(\tilde{\boldsymbol{y}}'_l - \tilde{\boldsymbol{y}}'_r)[2]$ (see Equation~\eqref{eq:epi}), with the keypoint RMSE.

\vspace{-5pt}
\subsection{Validation of 2D Certificates}
~\autoref{fig:2dcert_rmse} validates the effectiveness of the 2D certificate $\mathcal{OC}_{2D}$. This plot helps us compare how the IoU score, which can be computed at test time, correlates with the RMSE with the ground-truth keypoints.  A clear trend is observed: higher IoU scores correspond to lower RMSE values. Notably, pseudo-labels with IoU values exceeding 0.95 yield average keypoint errors below 10 pixels, which is small relative to the image resolution ($1640 \times 1232$). This empirical relationship supports the use of $\mathcal{OC}_{2D}$ as a reliable proxy for keypoint accuracy during pseudo-label validation.

\begin{figure}[t]
    \centering
    \begin{subfigure}[t]{0.31\textwidth}
        \centering
        \includegraphics[width=\linewidth]{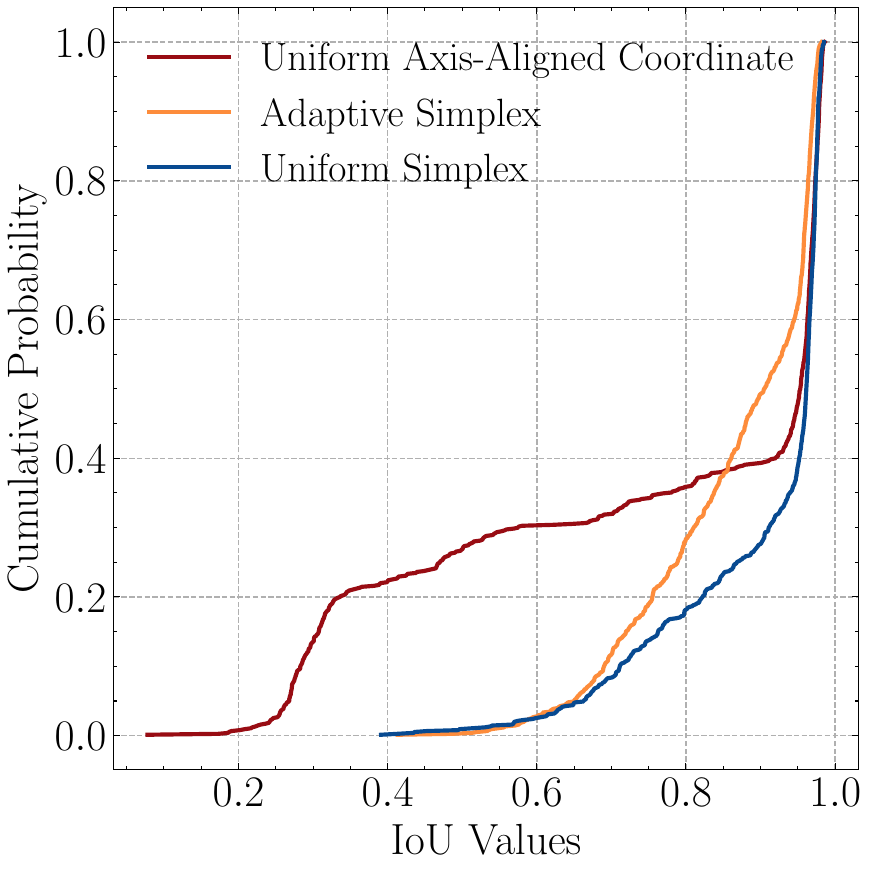}
        \caption{}
        \label{fig:sampling}
    \end{subfigure}
    \hfill
    \begin{subfigure}[t]{0.66\textwidth}
        \centering
        \includegraphics[width=\linewidth]{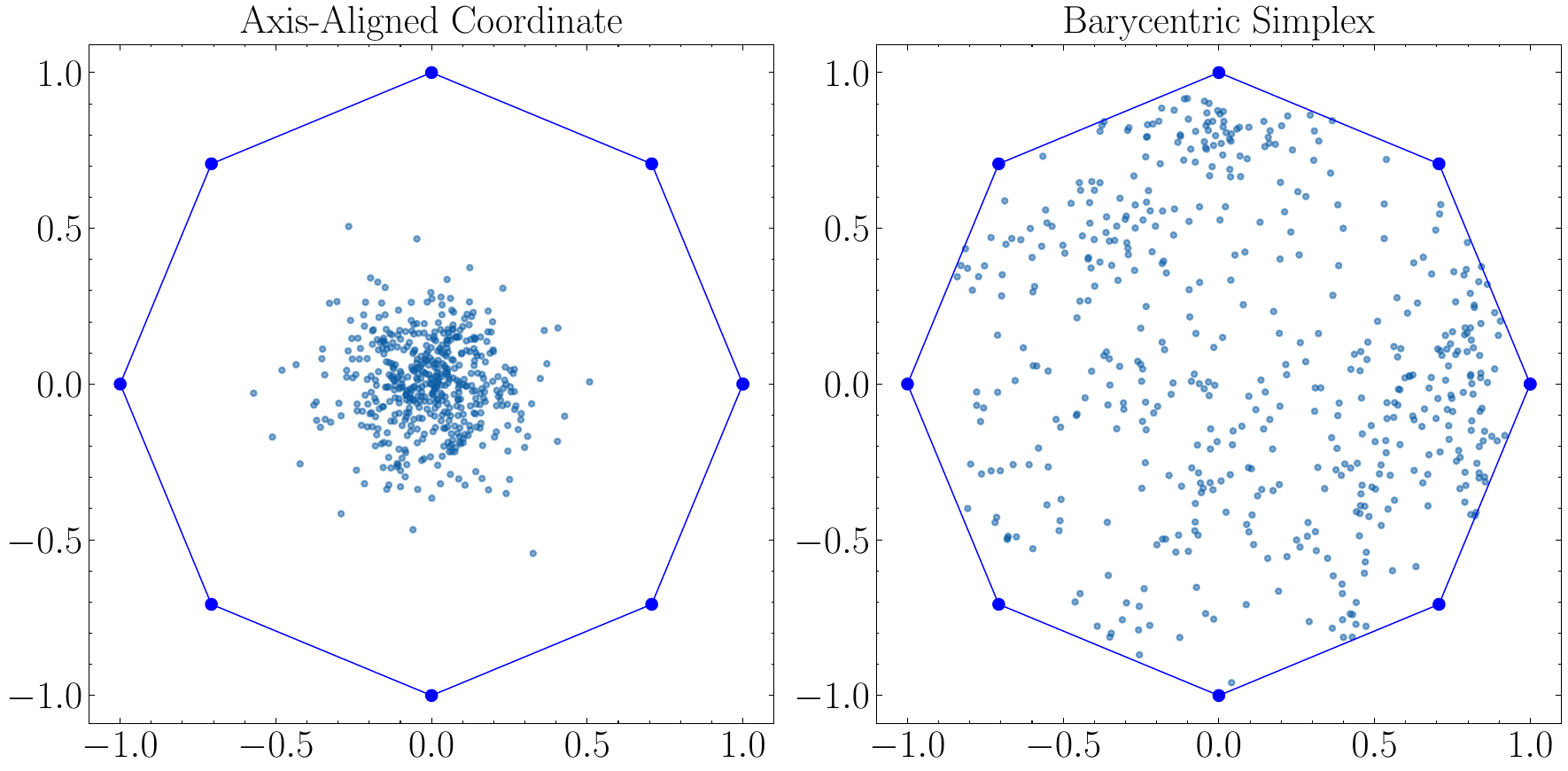}
        \caption{}
        \label{fig:simulate_sampling}
    \end{subfigure}
    \caption{Analysis of SAM2 sampling strategies. (a) Cumulative distribution of IoU values for three SAM2 sampling strategies. The x-axis shows the IoU between predicted and ground-truth segmentations, and the y-axis indicates cumulative probability. Uniform Simplex method outperforms both Adaptive Simplex and Uniform Axis-Aligned Coordinate. (b) Visual comparison of two sampling strategies within a regular octagon. With the same number of samples, the Axis-Aligned Coordinate sampling is densely concentrated near the center, while the Barycentric Simplex sampling provides more uniform coverage of the polygon.}
    \label{fig:combined_sampling}
\end{figure}

\vspace{-5pt}
\subsection{Validation of Residual Certificates}

\begin{figure}[t]
    \centering
    \begin{subfigure}[t]{0.32\textwidth}
        \centering
        \includegraphics[width=\linewidth]{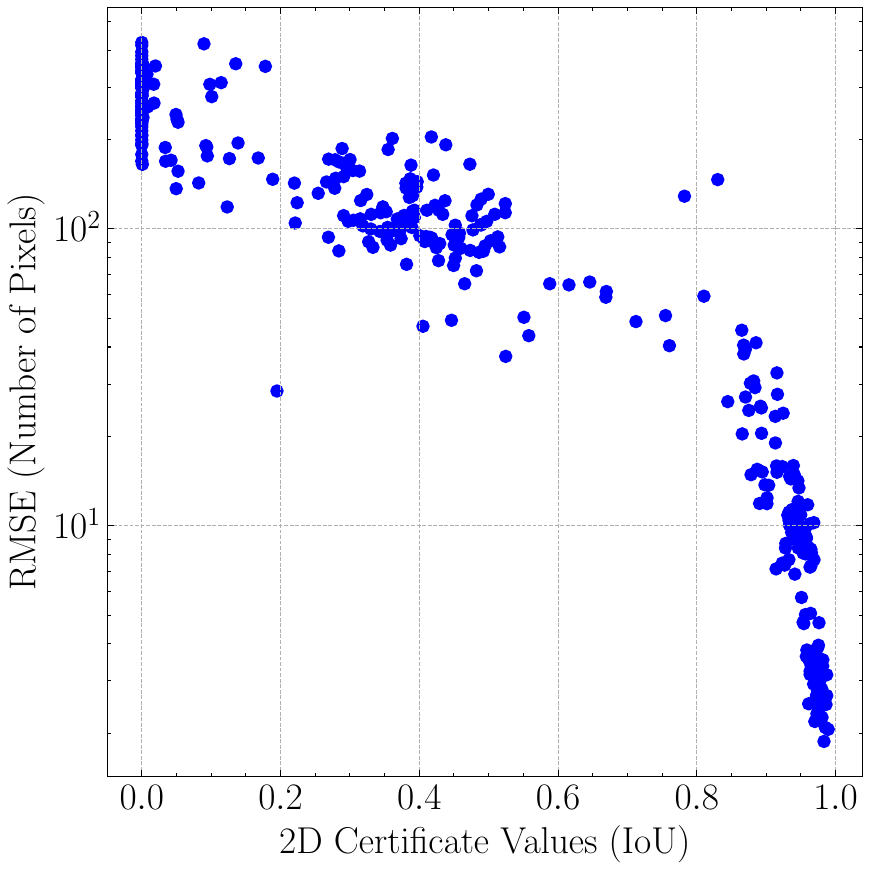}
        \caption{}
        \label{fig:2dcert_rmse}
    \end{subfigure}
    \hfill
    \begin{subfigure}[t]{0.32\textwidth}
        \centering
        \includegraphics[width=\linewidth]{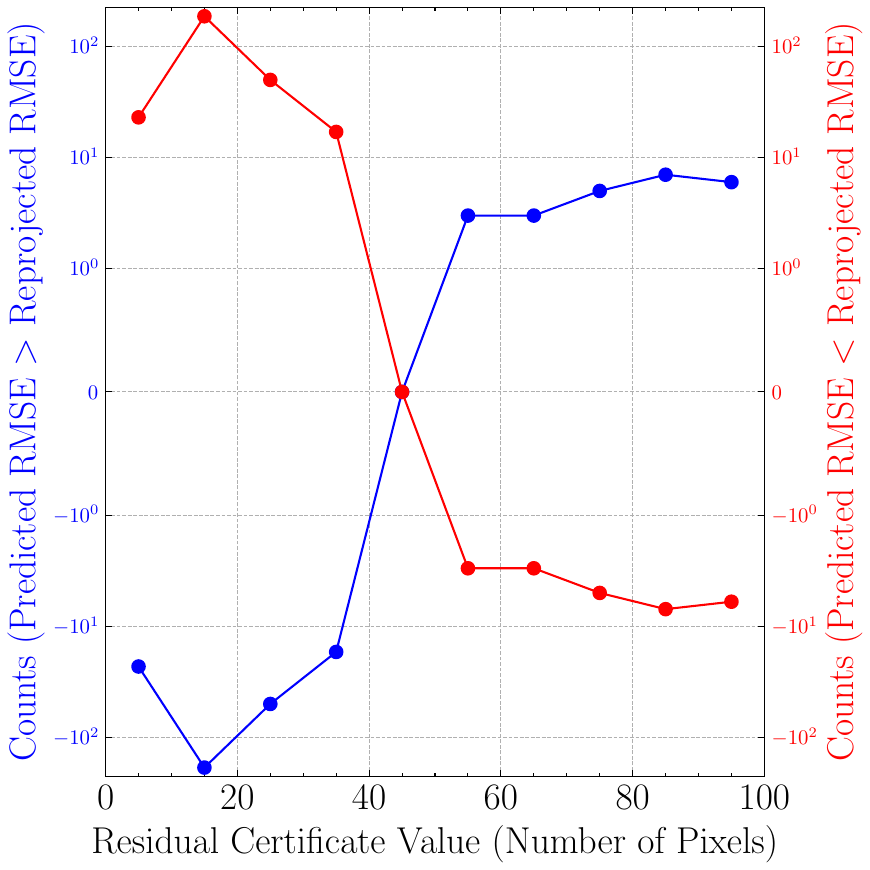}
        \caption{}
        \label{fig:res_cert}
    \end{subfigure}
    \hfill
    \begin{subfigure}[t]{0.32\textwidth}
        \centering
        \includegraphics[width=\linewidth]{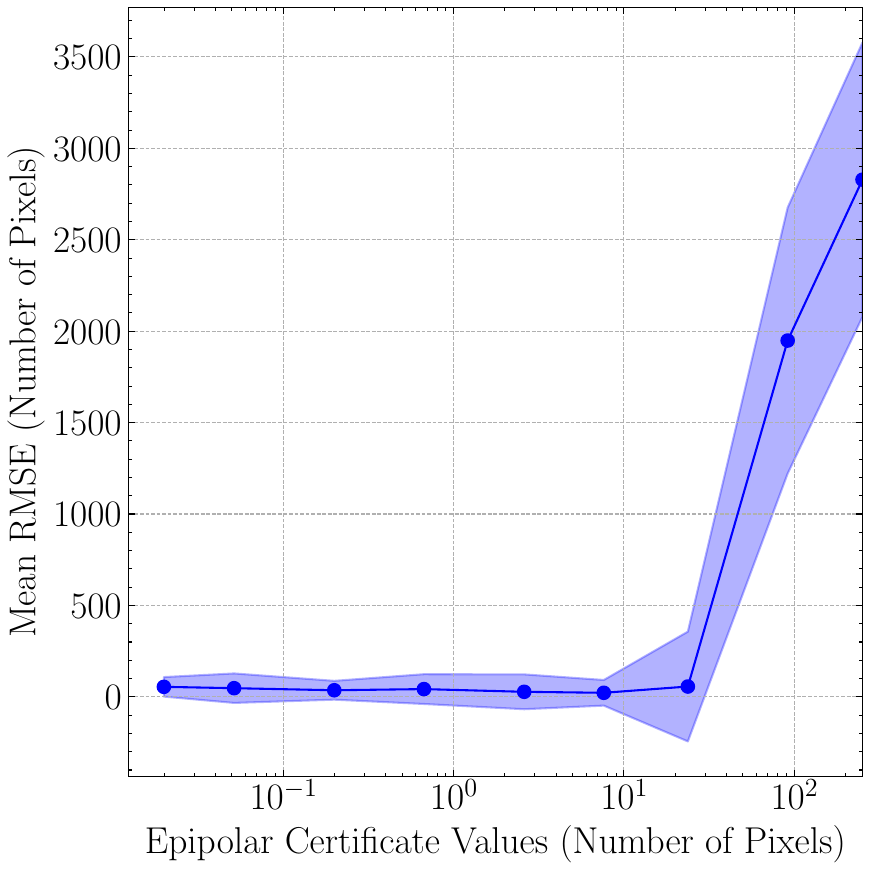}
        \caption{}
        \label{fig:epi_cert}
    \end{subfigure}
    \caption{(a)Validation of the 2D Certificate (\ie Equation~\eqref{eq:2d}). The x-axis is the IoU score in the $\mathcal{OC}_{2D}$ certificate. The y-axis is the RMSE error of the pseudo-labels averaged across one image sample. We use ground-truth segmentation for IoU calculation. Pseudo-labels with IoU values larger than 0.95 have average keypoint errors of fewer than 10 pixels which is small relative to the image size ($1640 \times 1232$). (b) Validation of the Residual Certificate (\ie Equation~\eqref{eq:res}). The x-axis is the residual value and the y-axis is the counts of keypoints that either predicted ones are more accurate (red) or reprojected ones are more accurate (blue). We found that there is a heuristic threshold that enables hybrid keypoint selection. (c) Validation of Epipolar Constraint Certificate (\ie Equation~\eqref{eq:epi}). The x-axis is the discrepancy of the y coordinates between rectified selected pseudo-label keypoints and ground truth keypoints in the $\mathcal{OC}_{epi}$ certificate. Y-axis is the RMSE error of the pseudo-labels averaged across one bin batch. This figure highlights the importance of enforcing epipolar constraint certificates.}
    \label{fig:cert_analy}
\end{figure}

In~\autoref{fig:res_cert} we validate the residual certificate $\mathcal{OC}_{res}$. We plot the residual certificate value $\left\|\boldsymbol{\delta} \right\|_2$ (see Equation~\eqref{eq:res}) on the x-axis. On the y-axis, we plot the count of instances where the predicted RMSE is either greater than (blue curve) or less than (red curve) the reprojected RMSE, across varying residual certificate values. The predicted RMSE is the RMSE of the predicted keypoints that are output directly from the keypoint network, and the reprojected RMSE is the RMSE of the optimized keypoints reprojected from the PnP pose and shape estimator. Note that the x-axis is the certificate value that can be computed at test time, whereas the y-axis (\ie predicted and reprojected RMSE) requires knowledge of the ground truth. 

We again observe a clear trend. For low residual values (left side of the x-axis), the majority of instances fall under the red curve, indicating that predicted keypoints are more accurate than reprojected ones. As the residual certificate increases, the trend reverses—beyond a residual value of approximately 42 pixels, the reprojected keypoints tend to outperform the predicted ones, as indicated by the rising blue curve. This transition point around 42 suggests an empirical threshold at which the residual certificate reliably filters high-quality predictions.

\vspace{-10pt}
\subsection{Validation of Epipolar Constraint Certificates}
In~\autoref{fig:epi_cert}, we validate the epipolar constraint certificates. We plot the epipolar certificate value (\ie Equation~\eqref{eq:epi})  on the x-axis. The corresponding mean RMSE of the selected pseudo-label keypoints and ground-truth keypoints (in pixels) are plotted on the y-axis. Note that while the epipolar certificate value can be computed at test time without the knowledge of the ground truth, the mean RMSE requires ground truth. We again observe a clear trend. As shown in~\autoref{fig:epi_cert}, we report the mean RMSE (in pixels) across test samples as a function of the epipolar certificate values. At low thresholds (\eg <20 pixels), the RMSE remains consistently low. However, as the value increases beyond 20 pixels, the RMSE grows rapidly, along with its variance. This behavior highlights the importance of enforcing epipolar certificates.

\vspace{-10pt}
\subsection{Impact of Sampling Strategies}
In scenarios where the ground truth mask $\boldsymbol{M}$ is unavailable—commonly the case in industrial settings—we leverage the SAM2 model~\cite{ravi2024sam2} to generate pseudo-ground truth masks for object boxes. To produce masks, SAM2 requires samples in the pixel space. In this section, we examine how various sampling strategies influence the quality of the resulting pseudo-ground truth masks.

~\autoref{fig:combined_sampling} analyzes the effect of different sampling strategies on SAM2 segmentations for a 2D convex polygon. We only discuss 2D convex polygon because the 2D projection of a box is a polygon. We consider three strategies:
\textbf{(1) Uniform Axis-Aligned Coordinate: }Candidates are generated by taking convex combinations of the polygon’s vertices. Specifically, we sample a non-negative weight for each vertex from a uniform distribution over $\left[0,1\right]$, then normalize the weights so that they sum to 1. \textbf{(2) Uniform Simplex:} Candidates are sampled uniformly from the convex hull of the polygon’s vertices using a triangulation-based approach. The polygon is first decomposed into simplices (i.e., triangles in 2D), and a simplex is selected via importance sampling, with the selection probability proportional to its area. A point is then sampled uniformly within the chosen simplex using barycentric coordinates, ensuring uniform coverage across the entire polygon~\cite{pharr2023physically}. \textbf{(3) Adaptive Simplex:} Similar to \textbf{Uniform Simplex}, but with a key difference: while \textbf{Uniform Simplex} uses a constant number of samples regardless of the area of the triangle, \textbf{Adaptive Simplex} scales the number of samples proportionally to the triangle’s area.

In \autoref{fig:sampling}, the cumulative IoU distribution shows that Uniform Simplex sampling significantly outperforms both Uniform Axis-Aligned Coordinate and Adaptive Simplex, achieving a higher proportion of accurate segmentation masks. \autoref{fig:simulate_sampling} visualizes the core difference by simulating sampling in a regular octagon. In Axis-Aligned Coordinate sampling, points tend to cluster densely near the center of the feasible region and are sparsely distributed near its boundaries while Simplex sampling generates points uniformly in the polygon.

\section{Experiments}
\label{sec:experiments}
We conducted three sets of experiments to evaluate \texttt{BOSS}. First, we validated the effectiveness of our pipeline on a synthetic dataset (Section \ref{sec:sim2sim}). Next, we demonstrated its ability to bridge the sim-to-real gap (Section \ref{sec:sim2real}). Finally, we will demonstrate its ability to perform self-supervised learning using a large-scale unlabeled dataset (Section \ref{sec:real2real}).

\begin{figure}[t]
    \centering
    \begin{subfigure}{0.32\columnwidth}
        \centering
        \includegraphics[width=\linewidth]{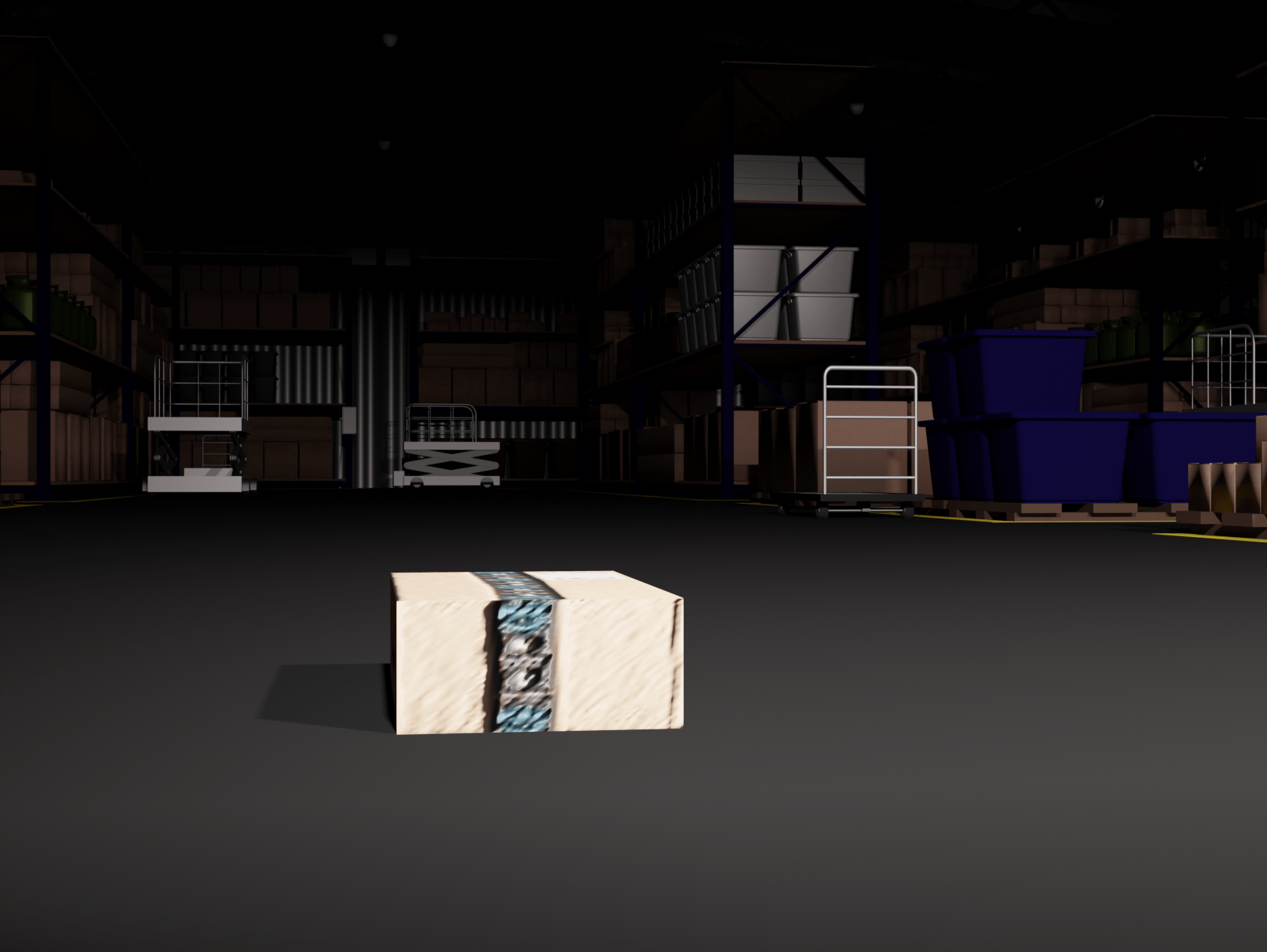}
        \caption{Simulation}
        \label{fig:simulation}
    \end{subfigure}
    \hfill
    \begin{subfigure}{0.32\columnwidth}
        \centering
        \includegraphics[width=\linewidth]{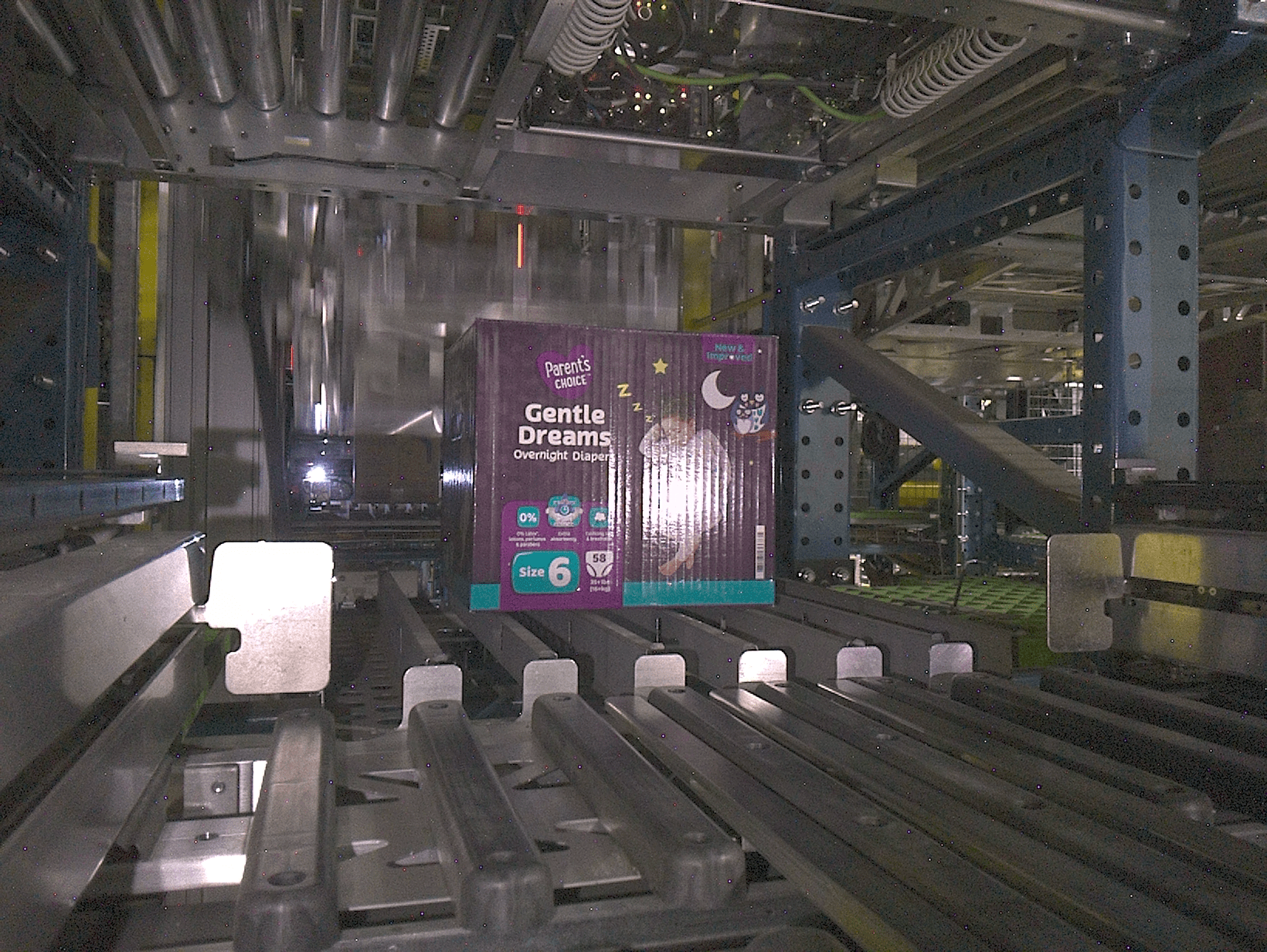}
        \caption{inbound buffer}
        \label{fig:inbound}
    \end{subfigure}
    \hfill
    \begin{subfigure}{0.32\columnwidth}
        \centering
        \includegraphics[width=\linewidth]{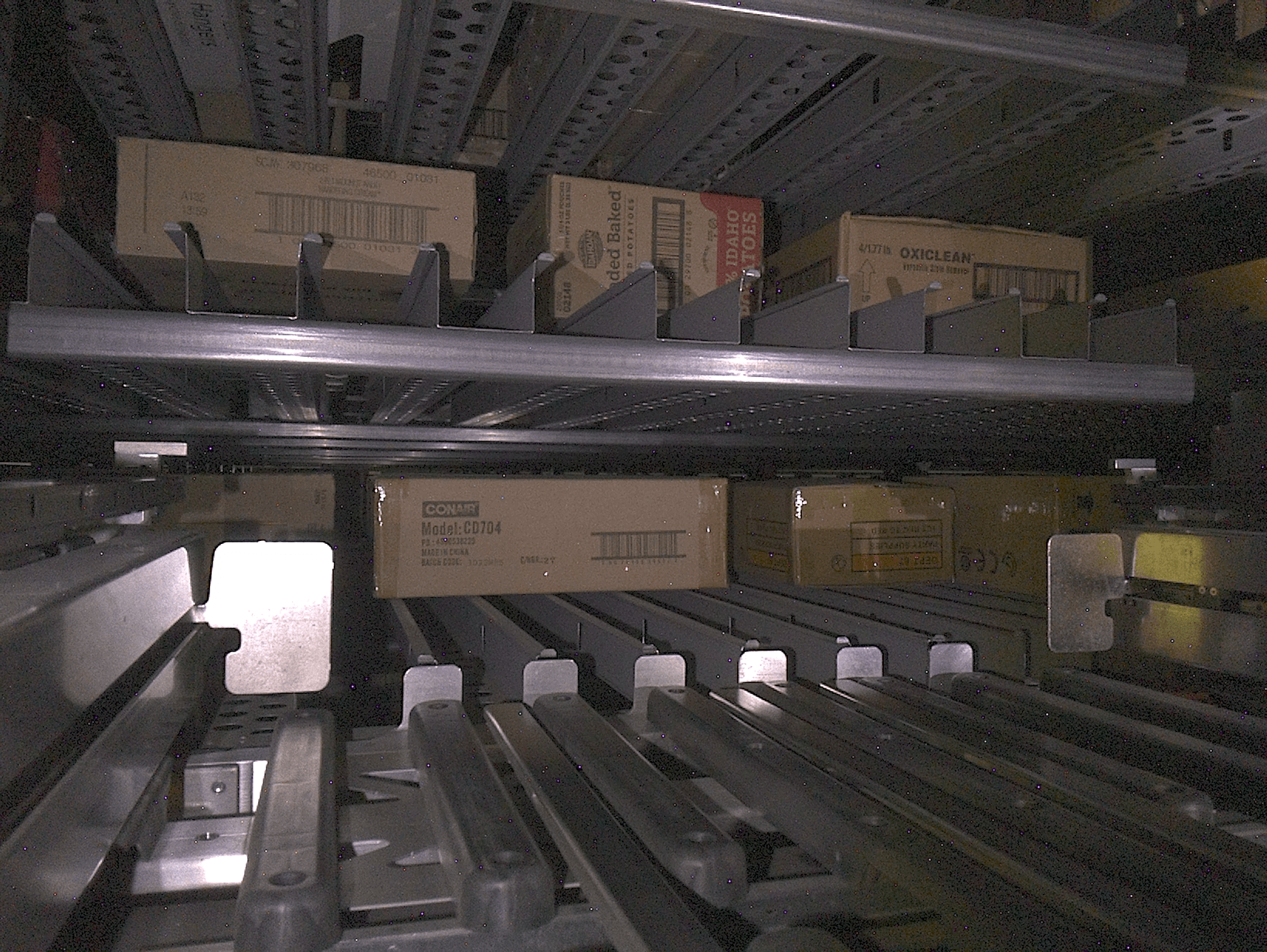}
        \caption{storage aisles}
        \label{fig:storage_aisles}
    \end{subfigure}
    \caption{Sample images from the simulated and real datasets used in the experiments.}
    \label{fig:combined_certificates}
\end{figure}

\begin{table}[h!]
  \centering
  \scriptsize
  \begin{tabularx}{\columnwidth}{l *{6}{Y}}
    \toprule
    & \multicolumn{6}{c}{\specialcell[b]{Pose and Shape Estimation Comparison}} \\
    \cmidrule{2-7}
    & \multicolumn{3}{c}{\textbf{Sim2Sim}} & \multicolumn{3}{c}{\textbf{Sim2Real}} \\
Approach & APE [m] & ARE [rad] & ASE [m] & APE [m] & ARE [rad] & ASE [m]\\
 \midrule 
\texttt{Model w/o SSL}  & 0.584  & 0.219  & 0.369 & 2.080       & 0.554    & 1.589      \\
\texttt{BOSS-SAM2}  & 0.038  & 0.069 & 0.084      & 0.134 & 0.223    & 0.238    \\
\texttt{BOSS-GT}  & 0.041  & 0.063  & 0.078     & 0.148       & 0.239       & 0.259  \\
\texttt{BOSS-SAM2 (50k)}  & -  &  - &   -  & 0.135       & 0.217      & 0.247  \\
  \midrule 
 \texttt{Model Supervised} & 0.024  & 0.053    & 0.045   & 0.111       & 0.212   & 0.208   \\
 
 \bottomrule
  \end{tabularx}%
  \caption{Pose and shape estimation for self-supervised pipeline and other baselines. APE denotes the average position error; ARE denotes the average rotation error; ASE denotes the average shape error.} 
  \label{tab:competitor_lc}
\end{table}{}

\subsection{Validation on Synthetic Dataset}
\label{sec:sim2sim}
\begin{figure}[t]
    \centering
    \begin{subfigure}{0.45\textwidth}
        \centering
        \includegraphics[width=\linewidth]{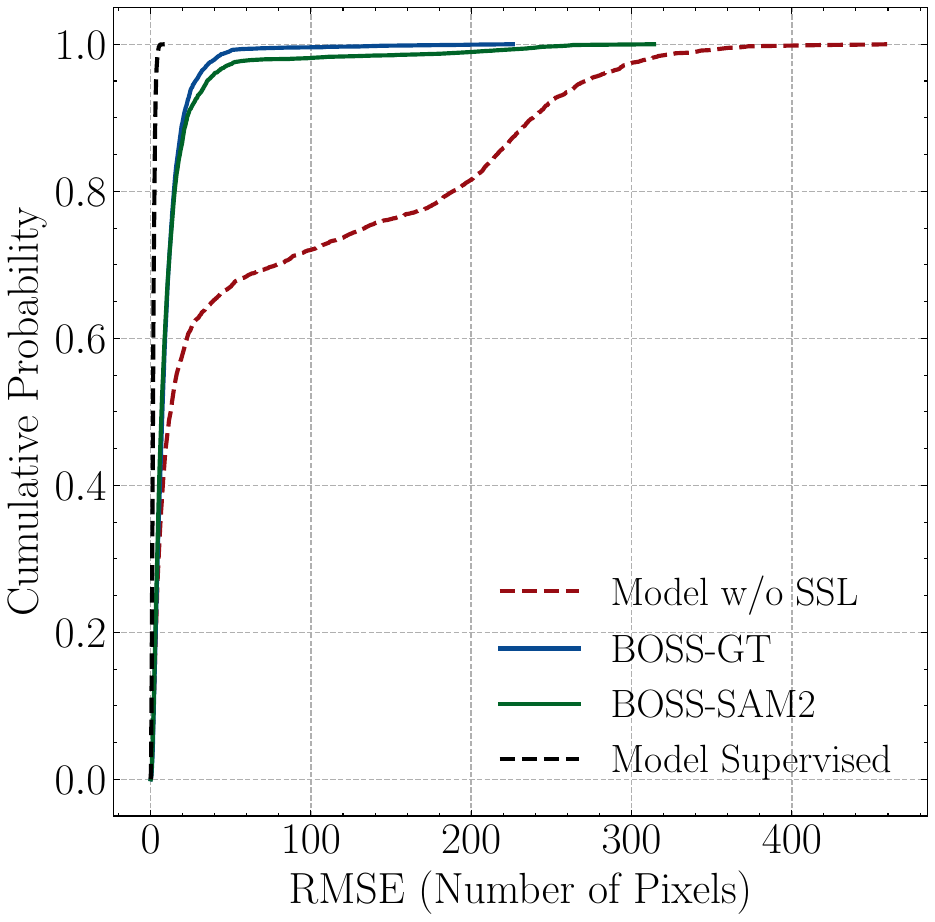}
        \caption{Sim2Sim keypoint detection comparisons for the proposed self-supervised architecture with upper bound and other baselines.}
        \label{fig:sim2sim}
    \end{subfigure}
    \hfill
    \begin{subfigure}{0.45\textwidth}
        \centering
        \includegraphics[width=\linewidth]{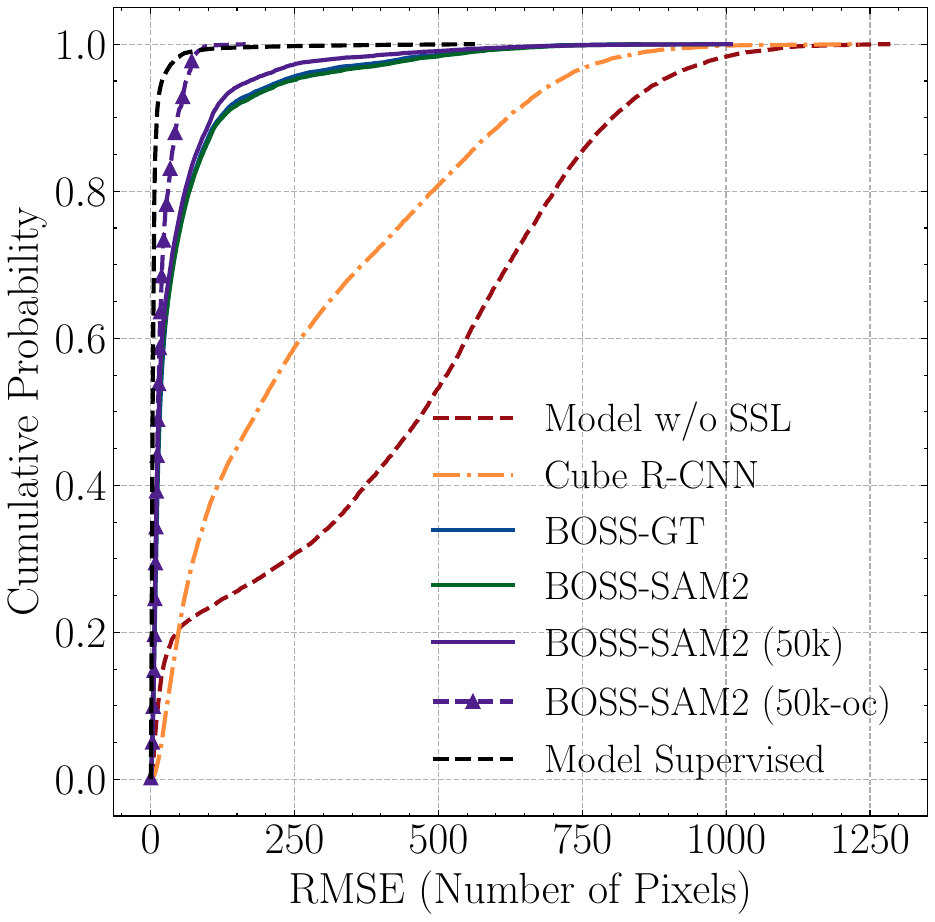}
        \caption{Sim2Real keypoint detection comparisons for the proposed self-supervised architecture with zero-shot large model and other baselines.}
        \label{fig:sim2real}
    \end{subfigure}
    \caption{Comparison of keypoint detection performance in Sim2Sim (left) and Sim2Real (right) scenarios.}
    \label{fig:sim2sim2real}
\end{figure}

 \textbf{Setup.} We use Blender to generate a dataset comprising a training dataset of 75 images and a test dataset of 375 images featuring five types of boxes. A typical example from the synthetic dataset is shown in~\autoref{fig:simulation}. The training dataset includes images captured from a fixed viewpoint of a single object type with varying lighting conditions and randomized object poses, while the test dataset features both novel views of known objects and entirely new objects. We test \texttt{BOSS}' ability to perform self-supervised learning on the test dataset. \\
\textbf{Results and Insights.} Keypoint detection results are shown in~\autoref{fig:sim2sim}. The baseline model without self-supervised learning \texttt{Model w/o SSL} is trained solely on the simulation training dataset. In contrast, the self-supervised models are trained on the same dataset but also perform self-supervised learning on the test dataset without annotations. This model has two variations: one using ground truth segmentation for the 2D certificate \texttt{BOSS-GT} and another using SAM2 masks \cite{ravi2024sam2} \texttt{BOSS-SAM2}. Finally, the supervised model \texttt{Model Supervised} is trained directly on the simulation test dataset (\ie this is the best achievable performance with the architecture). Our goal is to fill the area between curves of \texttt{Model w/o SSL} and \texttt{Model Supervised}, commonly known as the domain gap. Notably, SSL effectively bridges this gap, with up to $90\%$ of keypoints exhibiting errors below 20 pixels—remarkably small relative to the image resolution of $1640 \times 1232$.~\autoref{tab:competitor_lc} presents the pose and shape statistics. The model with SSL significantly enhances pose and shape estimation, achieving accuracy more than 10 times higher ---nearly matching that of a supervised model--- with only around 4cm average error for position estimation; for reference, the average dimension of the simulated boxes is 0.23m. Notably, for both keypoint detection and pose and shape estimation, the SAM2 variant performs comparably to the GT variant.

\vspace{-5pt}
\subsection{Adaptation to Real Dataset}
\label{sec:sim2real}
\textbf{Setup.} Symbotic provided a dataset with 9,000 images (Symbotic-9k), including various types of boxes in two industrial environments: buffer shelves at inbound (\autoref{fig:inbound}) and storage aisles (\autoref{fig:storage_aisles}). The dataset provides keypoint annotations for stereo images, with keypoints predefined as the box corners. Symbotic-9k is split into 7k/0.5k/1.5k images for train/val/test respectively. \\
\textbf{Results and Insights.} We evaluate all models on the test dataset split and show results in~\autoref{fig:sim2real}. \texttt{Model w/o SSL}, trained solely on synthetic data, serves as a lower bound. The upper bound model \texttt{Model Supervised} is trained and validated on the train/val dataset. The area between \texttt{Model w/o SSL} and \texttt{Model Supervised} is referred to as a sim-to-real gap. \texttt{BOSS} has two variations using GT segmentation \texttt{BOSS-GT} or SAM2 \texttt{BOSS-SAM2}. Both models are first trained on synthetic data and then refined through self-supervised learning on the train and validation datasets. For comparison, we include \texttt{Cube R-CNN}, an RGB-only zero-shot bounding box prediction model trained on the large-scale Omni3D benchmark \cite{brazil2023omni3d} (234k images) using 48 V100 GPUs, covering both indoor and outdoor environments. The results clearly show that the SSL models, initially trained on synthetic data and adapted using GT or SAM2-learned segmentation, successfully bridge the sim-to-real gap. It also outperforms \texttt{Cube R-CNN} by a large margin.~\autoref{tab:competitor_lc} presents detailed results on the pose and shape estimation. Since ground truth pose and shape are unavailable for the real dataset, we generate pseudo ground truth by running our pose and shape estimator on ground truth keypoints. Notably, the self-supervised model consistently improves both pose and shape estimation with significantly lower errors for all position, rotation, and shape estimation. Self-supervised baseline also approaches the performance of the supervised upper bound. We also observe that, for both keypoint detection and the pose and shape estimation, the SAM2 variation has a very similar performance to the GT variation.

\vspace{-5pt}
\subsection{Adaptation to Large-scale Dataset}
\label{sec:real2real}
\textbf{Setup.} We are interested in how performance scales with the size of the dataset. Symbotic provides an additional dataset of about 50,000 images, referred to as Symbotic-50k, which however has no ground-truth keypoint annotations. \texttt{BOSS-SAM2 (50k)} is first pre-trained on synthetic data and then refined via self-supervised learning using a combination of the train and validation datasets, along with Symbotic-50k. Note that for all \texttt{BOSS-GT}, \texttt{BOSS-SAM2}, and \texttt{BOSS-SAM2 (50k)}, we use the same certificate thresholds to have a fair comparison. We additionally report the performance of the model when evaluated only on outputs that pass all certificate checks, denoted as \texttt{BOSS-SAM2 (50k-oc)}. We present the keypoint detection results on the test split in~\autoref{fig:sim2real}. Pose and shape estimation results are presented in~\autoref{tab:competitor_lc}.\\
\textbf{Results and Insights.} Interestingly, \texttt{BOSS-SAM2 (50k)} outperforms \texttt{BOSS-SAM2} and \texttt{BOSS-GT} by a small margin. This suggests that keypoint detection performance scales with dataset size. We can gain further performance improvement by filtering out bad labels during inference as shown \texttt{BOSS-SAM2 (50k-oc)}, whose performance is quite close to that of the supervised baseline. However, the improvement of \texttt{BOSS-SAM2 (50k)} compared with \texttt{BOSS-SAM2} is limited. We believe this can be improved in the future by an automatic certificate threshold update scheme during training. The current training uses a fixed threshold profile.

\section{Conclusions}
\label{sec:conclusions}

A self-supervised approach can train a box pose and shape estimation model using large-scale, unannotated data collated by a robot fleet in a warehouse.
Implementing a simple pipeline to estimate the pose and shape of a box, we show that it can be self-trained leveraging our correct-and-certify approach. 
The correct-and-certify approach implements certificates to pseudo-label instances during training but requires hard thresholds to be set apriori for training. We devise an empirical way to choose these thresholds and demonstrate that our training can bridge a large domain gap. 
Several avenues remain open for future research. First, rather than applying hard thresholds to model outputs, can we use soft pseudo-labels to retain more information? This idea is motivated by the observation that certificate values naturally reflect the confidence level of each pseudo-label. Second, we are interested in extending pose and shape estimation to irregularly shaped objects, which would significantly improve generalization across diverse warehouse tasks. Potential solutions include incorporating shape parametrization \cite{shan2021ellipsdf} or learning a latent shape representation \cite{mittal2022autosdf,liu2023openshape}.

%
%
%

\bibliographystyle{IEEEtran}
\bibliography{main}







\end{document}